\documentclass{IEEE_lsens}
\usepackage{textcomp}

\usepackage[noadjust]{cite}

\ifCLASSINFOpdf
\else
\fi
\usepackage[T1]{fontenc} 
\usepackage{amsmath}
\usepackage{graphicx}%
\usepackage[cmintegrals]{newtxmath}
\usepackage{bm}

\usepackage{array}
\usepackage{url}
\ifCLASSINFOpdf
\else
\fi
\providecommand{\hypersetup}[1]{\relax}

\hypersetup{pdftitle={Bare Demo of IEEE\_lsens.cls for IEEE Sensors Letters},
pdfsubject={Typesetting},
pdfauthor={Michael D. Shell},
pdfkeywords={Class, IEEE, IEEE\_lsens, IEEE Sensors Letters, LaTeX, Typesetting, TeX}}
\begin{document}

\markboth{Vol.~1, No.~3, July~2017}{0000000}

\IEEELSENSarticlesubject{Sensor Applications}

%
\title{LG-Sleep: Local and Global Temporal Dependencies for Mice Sleep Scoring}

%
\author{\IEEEauthorblockN{Shadi~Sartipi\IEEEauthorrefmark{1}, Mie~Andersen\IEEEauthorrefmark{2},Natalie~Hauglund\IEEEauthorrefmark{2},Celia~Kjaerby\IEEEauthorrefmark{2},Verena~Untiet\IEEEauthorrefmark{2},Maiken~Nedergaard\IEEEauthorrefmark{2},
and~Mujdat~Cetin\IEEEauthorrefmark{1,3}\IEEEauthorieeemembermark{1}}
\IEEEauthorblockA{\IEEEauthorrefmark{1}Department of Electrical and Computer Engineering,
University of Rochester, Rochester, NY, USA\\
\IEEEauthorrefmark{2}Division of Glial Disease and Therapeutics, Center for Translational Neuromedicine, University of Copenhagen, Copenhagen, Denmark\\
\IEEEauthorrefmark{3}Goergen Institute for Data Science,
University of Rochester, Rochester, NY, USA\\
\IEEEauthorieeemembermark{1}Fellow, IEEE}%
\thanks{Corresponding author: S. Sartipi (e-mail: ssartipi@ur.rochester.edu).\protect\\
This work has been partially supported by the National Science Foundation (NSF) under grants CCF-1934962 and DGE-1922591, and by the BRAIN Initiative of the National Institutes of Health through grant U19NS128613.}}
%
%
%


\IEEEtitleabstractindextext{%
\begin{abstract}
Efficiently identifying sleep stages is crucial for unraveling the intricacies of sleep in both preclinical and clinical research. The labor-intensive nature of manual sleep scoring, demanding substantial expertise, has prompted a surge of interest in automated alternatives. Sleep studies in mice play a significant role in understanding sleep patterns and disorders and underscore the need for robust scoring methodologies. In response, this study introduces LG-Sleep, a novel subject-independent deep neural network architecture designed for mice sleep scoring through electroencephalogram (EEG) signals. LG-Sleep extracts local and global temporal transitions within EEG signals to categorize sleep data into three stages: wake, rapid eye movement (REM) sleep, and non-rapid eye movement (NREM) sleep. The model leverages local and global temporal information by employing time-distributed convolutional neural networks to discern local temporal transitions in EEG data. Subsequently, features derived from the convolutional filters traverse long short-term memory blocks, capturing global transitions over extended periods. Crucially, the model is optimized in an autoencoder-decoder fashion, facilitating generalization across distinct subjects and adapting to limited training samples. Experimental findings demonstrate superior performance of LG-Sleep compared to conventional deep neural networks. Moreover, the model exhibits good performance across different sleep stages even when tasked with scoring based on limited training samples.
\end{abstract}

}


\maketitle

\section{Introduction}\label{sec1}

The study of sleep has gained widespread interest due to its impact on personal well-being, cognitive performance, and memory function. It is crucial to understand the sleep stages and transitions between stages to explain how the brain functions and recovers during rest. One common approach to recognizing these stages is analyzing the electrical brain activity patterns while the subject is asleep using electroencephalogram (EEG) measurements \cite{wei2018research}. By examining the characteristic waveforms and frequencies present in EEG signals, researchers can identify and differentiate various sleep stages, such as wakefulness, rapid eye movement (REM) sleep, and non-rapid eye movement (NREM) sleep \cite{fisher2012rapid}. This analysis not only contributes to our comprehension of sleep architecture but also offers significant implications for the diagnosis and treatment of sleep disorders. 

Manual analysis and inspection by human experts is the traditional way to detect sleep stages. However, this process can be time-consuming and requires a high level of expertise, which negatively affects the scalability of subjects and the number of recordings \cite{stephenson2009automated,barger2019robust}. As a solution to this challenge, automated scoring systems have been proposed, which have been extensively studied since the late 1960s \cite{smith1969automatic, drane1969pattern}. By automating the scoring process, these systems can reduce human error rates, increase accuracy, and lower the costs \cite{barger2019robust}. 

Animal studies are essential in sleep research, providing significant insights into the mechanisms governing sleep and sleep-related disorders \cite{paterson2011sleep}. Traditional approaches for automating sleep scoring in rodents depend on hand-crafted features, reflecting expert interpretations of sleep data. Common features include power spectral density (PSD), wavelet coefficients, and EEG data amplitude \cite{paterson2011sleep, brankavck2010eeg}. These features capture the frequency and statistical aspects of EEG data \cite{brankavck2010eeg, grieger2021automated}. Although effective, the challenge with hand-crafted features lies in the exhaustive computation required for all possible features across time and frequency domains. This process can be both complex and time-intensive given the noisy nature of EEG data.

The use of machine learning, specifically deep learning-based approaches within the domain of artificial intelligence, is gaining popularity in the medical field \cite{yamabe2019mc, alsolai2022employing}. Although deep learning research in human sleep scoring produced many methods and articles\cite{fiorillo2019automated}, studies conducted on rodents are limited \cite{barger2019robust}. Convolutional neural networks (CNNs) \cite{sartipi2021eeg,kiranyaz2015real} and long short-term memory (LSTM) models have been commonly used in the context of EEG data analysis \cite{graves2005framewise}. CNNs extract effective features with convolutional filters from images and have also been studied in physiological data analysis \cite{anwar2018medical}. While CNNs have demonstrated excellent performance in identifying frequency-based features, they often fall short in capturing temporal characteristics. To address this limitation, LSTM is employed to encode the transition rules derived from long time-series data. Despite deep learning's effectiveness in numerous disciplines, the imbalance problem in sleep stages (wake/REM/NREM) challenges the model's performance \cite{soltani2019sleep}. Another challenge is individual differences, which can decrease the learned model's performance on new subjects' data.

To address the challenges of the imbalance problem in sleep stages and individual differences, we propose a deep neural network architecture called LG-Sleep. This architecture takes into account both local and global temporal dependencies. Given that manual mice sleep scoring is conducted in short intervals, incorporating information from adjacent stages could enhance the model's ability to detect valuable characteristics of the data. The backbone of our model incorporates CNN and LSTM to automatically extract features from EEG data. Time-distributed convolutional neural networks are applied to EEG data to extract the local temporal transitions. Then, the LSTM blocks are used to obtain the global temporal transitions for each data sample. To ensure that the model is robust against limitations in the number of training samples, we propose the model in the encoder-decoder fashion. One can think of the learned model as optimizing the discrepancies between the input data and the decoder's output while minimizing the classification error to ensure that the learned features are related to sleep stages. By extracting the local and global temporal dependencies, the model becomes more generalizable across subjects. 



\section{Materials and Method}
\label{sec:method}
\subsection{Dataset}
\subsubsection{Mice} Wildtype $C57BL/6$ mice were acquired from Janvier Labs at 7 weeks of age. Male and female mice were given unrestricted access to food and water with a normal $12$-hour light/dark cycle. The animals were $12-20$ weeks old at behavioral assessment. All experiments were approved by the Danish Animal Experiments Inspectorate and overseen by the University of Copenhagen Institutional Animal Care and Use Committee (IACUC), in compliance with European Communities Council Directive of 22 September 2010 (2010/63/EU) on animal protection in scientific research.

\subsubsection{Surgery} Mice underwent surgery between 7-15 weeks old, with anesthesia administered at $5\%$ isoflurane and maintained at $1-3\%$. The mice received preoperative buprenorphine for general analgesia along with lidocaine at the incision site. $0.8$ mm low impedance stainless steel screws (NeuroTek) were inserted into two burr holes above the frontal cortex (contralateral to the optic implant) and the cerebellum (reference area). Before waking up, animals received carprofen ($5$ mg/kg) s.c.. At least two weeks were allowed for recovery.

\subsubsection{Sleep measurements} Mice were placed in recording chambers (ViewPoint Behavior Technology), and EEG signals were amplified (National Instruments Inc., 16-channel AC amplifier Model 3500). In this work, only EEG signals were analyzed. EEG data were filtered with a passband of 1-100 Hz, and a 50 Hz notch filter reduced power line noise. Signals were digitized using a Multifunction I/O DAQ device and sampled at 512 Hz. Vigilance states were defined as wake (high muscle tonus, high frequency, low amplitude EEG), NREM sleep (no muscle tonus, low frequency, high amplitude EEG), and REM sleep (no muscle tonus, high frequency, low amplitude EEG). Details can be found in \cite{kjaerby2022memory}.

\begin{figure}[t!]
    \centering
    \includegraphics[width=\linewidth]{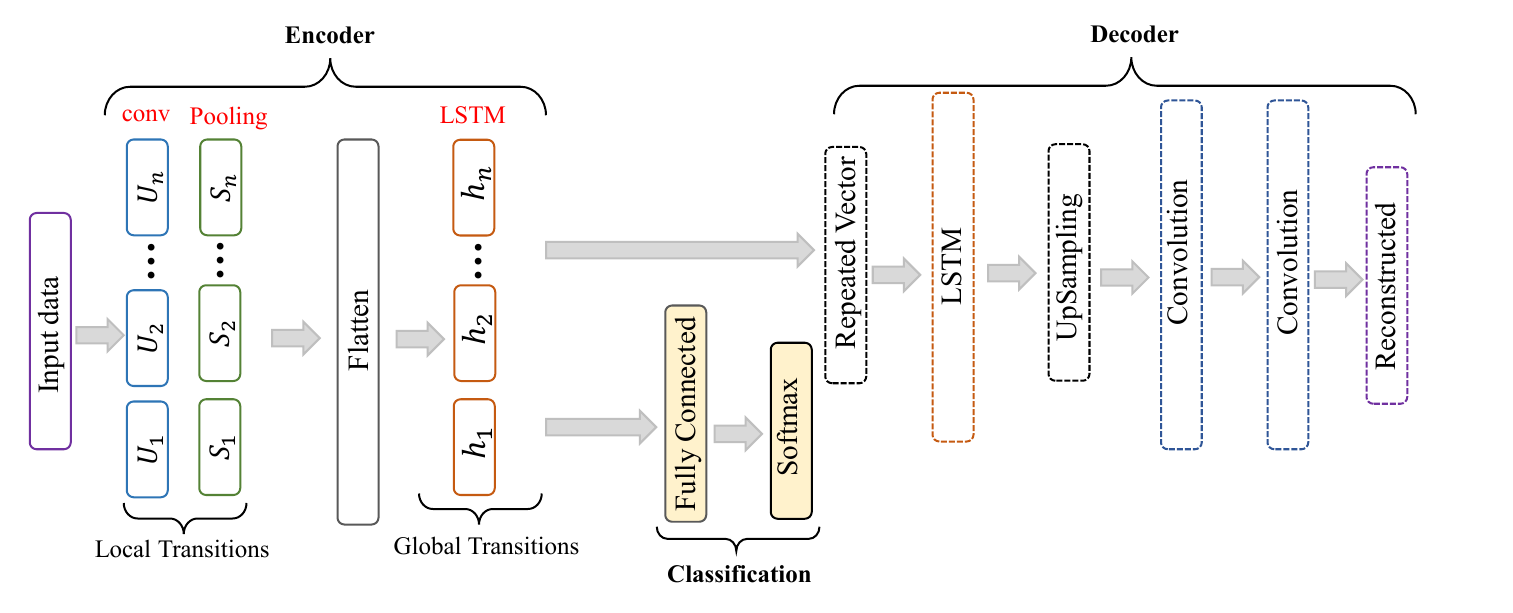}
    \caption{Overview of the proposed LG-Sleep deep neural network.}
    \label{fig: overall}
\end{figure}
\begin{table}[t!]
\centering
\caption{Parameters of the layers in LG-Sleep. Parameter format: CNN (number of filters; filter size) with ReLU activation, Pool (pool size), Dropout (dropout rate), LSTM (filter size, dropout rate)}
\vspace{-0.3 cm}
\begin{tabular}[t]{ c c |c c }
\hline
 \multicolumn{2}{c}{Encoder} & \multicolumn{2}{c}{Decoder} \\
\hline
\multicolumn{1}{c}{layer}&\multicolumn{1}{c|}{parameter} & \multicolumn{1}{c}{layer} &  \multicolumn{1}{c}{parameter} \\
\hline
CNN&$(64;64)$&LSTM&$(512;0.2)$\\
Pool&$2$&Reshape&$(19,8,64)$\\
Dropout&$0.5$&UpSample and CNN&$32$ and $(64;9)$\\
LSTM&$(32;0.4)$&UpSample and CNN &$2$ and $(1;1)$\\
\hline
\end{tabular}
\label{tab:detailsofmodel}
\end{table}
\begin{figure}[t!]
    \centering
    \begin{minipage}[c]{0.49\linewidth}
    \includegraphics[width=0.9\linewidth]{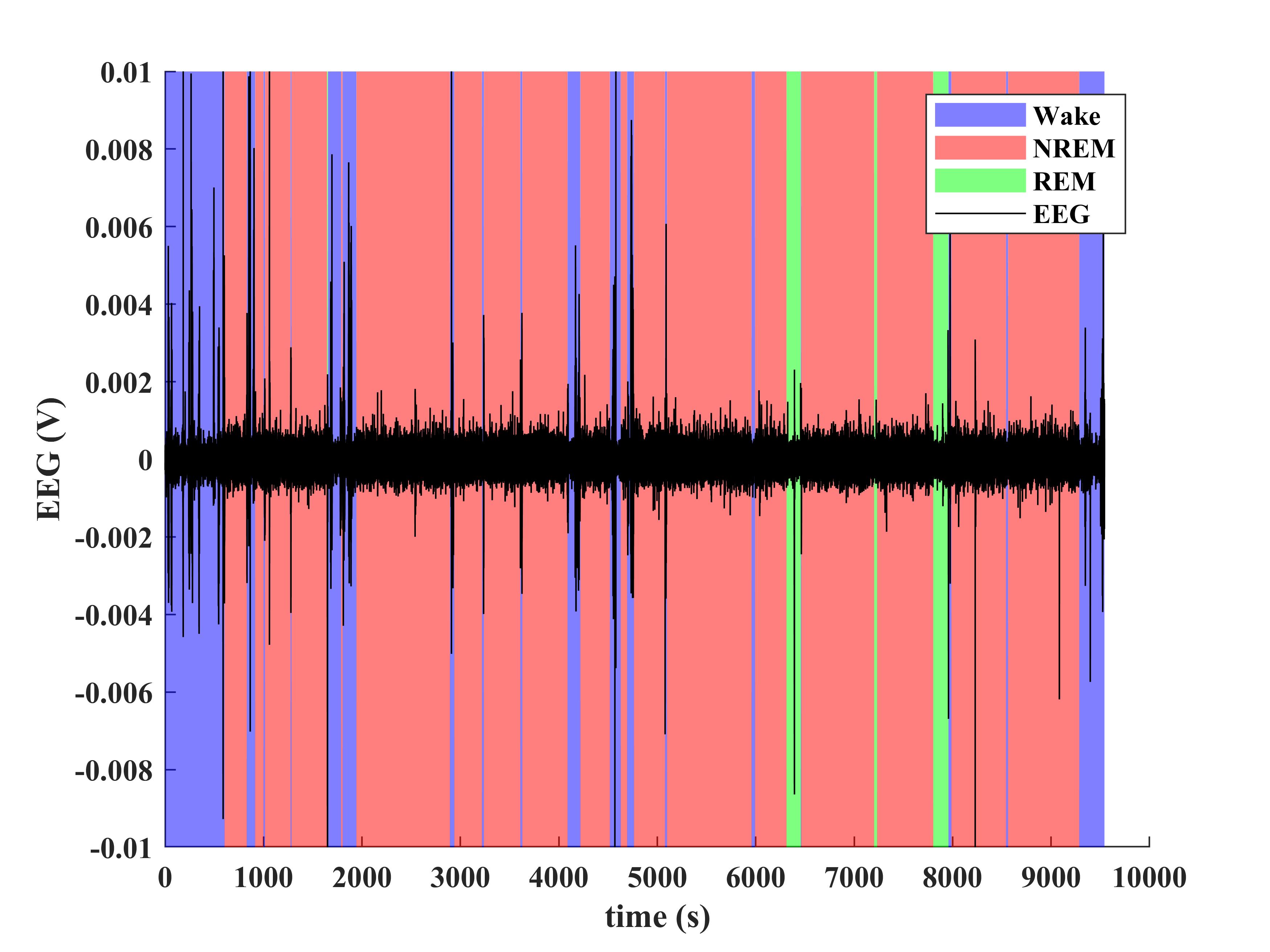}\\
    \centering{(a)}
    \end{minipage}
    \begin{minipage}[c]{0.49\linewidth}
    \includegraphics[width=0.9\linewidth]{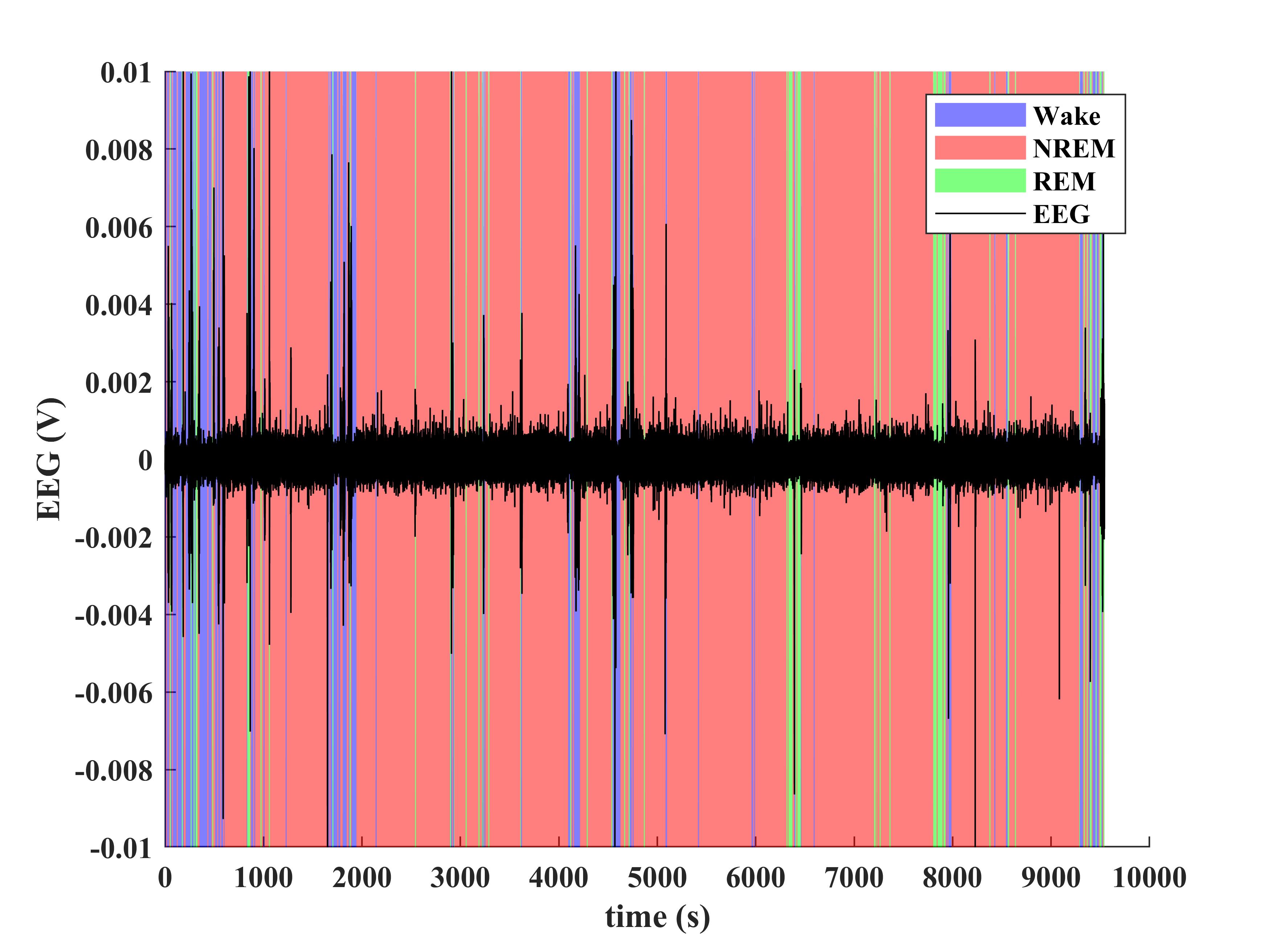}\\
    \centering{(b)}
    \end{minipage}    
    \caption{Sample EEG trace and the corresponding sleep stages: a) true labels, and b) predicted labels. The EEG trace is shown in black, and the background colors purple, red, and green correspond to the wake, NREM, and REM stages, respectively.}
    
    \label{fig:trace}
\end{figure}
\subsection{Data Preparation}
EEG data, $X$, are recorded over time, $T$, using electrodes to capture the brain's electrical activity. Due to susceptibility to noises like ocular artifacts and muscle movements, a bandpass filter ranging from $1$ to $45$ Hz is applied to cleanse the data. We use a slicing window of length $l$ to generate $N$ EEG trials, each segmented into $n$ slices with $1000$ ms duration and $500$ ms overlap. The LG-Sleep model processes these by first capturing local temporal dependencies within each segment, then addressing global dependencies every $l=10$ seconds.
\subsection{LG-Sleep}
LG-Sleep is composed of two main parts, the encoder and the decoder. The encoder component consists of a $1$D CNN layer that performs a convolution operation to capture effective local temporal representations. In this work, we set the width of the convolution kernel to $500$ ms to explore the temporal dynamics, and the filter size of the CNN layer is empirically set to $64$. The EEG temporal slices are encoded to a higher representation, denoted by $U_{i}$, where $i={1,2,\dots,n}$, using the rectified linear unit (ReLU) activation function. A max-pooling layer is then applied to obtain the final convolutional representation, $S_{i}$, where $i={1,2,\dots,n}$.

To capture the global temporal transitions, an LSTM block is employed after the CNN layer. The LSTM block includes four gates: the input gate that controls the value flow in a cell, the cell state that recalls the information, the forget gate that retains the required information, and the output gate that controls the extent of the value flow in the output. The output of the CNN layer is flattened and fed into the LSTM block with a hidden state size of $32$. The LSTM layer is followed by the ReLU activation function and dropout $0.4$.

For the decoder part, all layers align with the encoder. The latent representation vector is repeated $n$ times and passed through the LSTM layer with a hidden state size of $256$. The LSTM output is upsampled by a factor of $8$ and then passed through two $1$D CNN layers with filter sizes of $64$ and $1$ to reconstruct the input data. To optimize the encoding and decoding process, the mean-square-error loss function $\mathcal{L}{mse}$ is adopted as $\mathcal{L}{mse} = \frac{1}{N} \sum_{i=1}^{N}(X_i-\hat{X}_{i})^{2}$, where $X$ and $\hat{X}$ are the original input data and reconstructed output.

For the classification step, the latent representation vector, $A$, is fed into a fully connected layer with $32$ neurons. The resulting representations are then passed through a softmax classifier, $Z =softmax(WA + b)$, where $W$ and $b$ are weight and bias matrices, respectively. The categorical cross-entropy loss function, $\mathcal{L}_c$, is used to evaluate the final classification performance. The final loss function is the sum of the two loss functions, i.e., $\mathcal{L}=\mathcal{L}_{c}+\mathcal{L}_{mse}$. Fig.~\ref{fig: overall} shows the overall structure of the proposed LG-Sleep architecture.
\subsection{Implementation}
In this study, EEG data is segmented every $10$ seconds into trials, then sliced into one-second intervals with a $500$ ms overlap, capturing temporal details. With a $512$ Hz sampling rate, each sample size is $19\times512\times1$. To explore frequency impacts, the data are filtered into six subbands: delta ($1$-$4$ Hz), theta ($4$-$8$ Hz), alpha ($8$-$12$ Hz), beta ($12$-$29$ Hz), gamma ($30$-$45$ Hz), and wide-band ($1$-$45$ Hz). The filter is an IIR filter of order 5, using forward-backward filtering to eliminate phase distortion.

The details of the LG-Sleep blocks are in Table~\ref{tab:detailsofmodel}. All CNN, pooling, and upsampling layers are one-dimensional time-distributed layers. The encoder's output vector is repeated $18$ times before passing through the LSTM decoder layer. All CNN layers include a batch-normalization layer. LG-Sleep's learning is divided into two stages. Initially, the model is optimized with both loss functions, $\mathcal{L}=\mathcal{L}_{c}+\mathcal{L}_{mse}$, to generalize features and enhance robustness to limited training samples. The second stage focuses solely on $\mathcal{L}_{c}$, adjusting for class weights to address imbalanced class distributions, set to $1.5$, $1$, and $7$ for wake, REM, and NREM, respectively. Training uses the Adam optimizer with rates of $0.0001$ and $0.001$ for the first and second stages, over $20$ and $100$ epochs, respectively.
\section{Experiments and Discussion}
\label{sec:resultd}


\begin{table}[t]
    \centering
    \caption{Performance of the proposed LG-Sleep architecture in different frequency sub-bands.}
    \vspace{-0.3 cm}
    {\begin{tabular}[t]{c c c| c c c}
    \hline
{Subband} & {Accuracy} &  {F1-score}& {Subband} & {Accuracy} &  {F1-score}\\
        \hline
    Delta               & $0.75\pm0.05 $ & $0.60\pm0.07$&Beta & $0.74\pm0.09$ & $0.53\pm0.07$\\
    Theta               & $0.73\pm0.05 $ & $0.65\pm0.07$&Gamma& $0.62\pm0.08$ & $0.29\pm0.04$\\
    Alpha               & $0.72\pm0.08$ & $0.51\pm0.07$&Wide-band& $0.85\pm0.09$ & $0.75\pm0.10$\\
    \hline
    \end{tabular}}
    \label{tab: table1}
\end{table}

To evaluate our proposed method, we address a subject-independent 3-class classification problem involving different sleep stages: $34\%$  wake, $58\%$  NREM, and $8\%$  REM episodes. We employ $4$-fold subject-out cross-validation on $16$ mice, repeated $4$ times to ensure reliability and average the performance outcomes. In each validation, $12$ mice were used as the training set, and $4$ mice were used as the test set. During training, $10\%$  of samples are used as validation data, and the model parameters that yield the minimum validation loss are selected as the optimal model for testing. Given the data imbalance across classes, we assess performance using accuracy and the macro F1-score. The macro F1-score is calculated for each class and averaged, ensuring equal consideration of each class’s performance.


As mentioned before, the training process contains two phases. First, the model is trained for $20$ epochs with equal class weights by optimizing $\mathcal{L}_c$ and $\mathcal{L}_{mse}$. Second, the training continues for $100$ more epochs with different class weights by optimizing $\mathcal{L}_c$. Fig.~\ref{fig:trace} shows a sample of the true labels versus predicted ones for the wide-band frequency range. 
To show the role of different frequency subbands in sleep scoring via EEG signaling, Table~\ref{tab: table1} presents the results for each frequency subband by reporting the means and standard deviations of accuracies and F1-scores across all cross-validation folds. As expected, the results indicate that in the LG-Sleep architecture, the performance with gamma band data is lower than the other bands, which is consistent with the role of lower frequencies in sleep scoring \cite{benington1994scoring,huang2021eeg}. On the other hand, the architecture performs poorly on the gamma subband, with an accuracy of $0.62$ and an F1-score of $0.29$. The wide-band subband achieves the highest accuracy and F1-score among all the subbands, with an accuracy of $0.85$ and an F1-score of $0.75$. Overall, the results suggest the effectiveness of LG-Sleep. 

\begin{table}[t]
    \centering
    \caption{Ablation experiments to analyze different backbones.}
    \vspace{-0.3 cm}
    {\begin{tabular}[t]{c c c}
    \hline
\textbf{Method} & {Accuracy} &  {F1-score}\\
        \hline
    {\bf Proposed LG-Sleep}               & ${\bf 0.85\pm0.09}$ & ${\bf 0.75\pm0.10}$\\
    CNN & $0.73\pm0.09$ & $0.57\pm0.07$\\
    LSTM& $0.75\pm0.08$ & $0.61\pm0.10$\\
    FC& $0.66\pm0.08$ & $0.48\pm0.04$\\
    \hline
    \end{tabular}}
    \label{tab: ablation}
\end{table}
\begin{figure}[t]
    \centering
    \begin{minipage}[c]{0.49\linewidth}
    \includegraphics[width=0.75\linewidth]{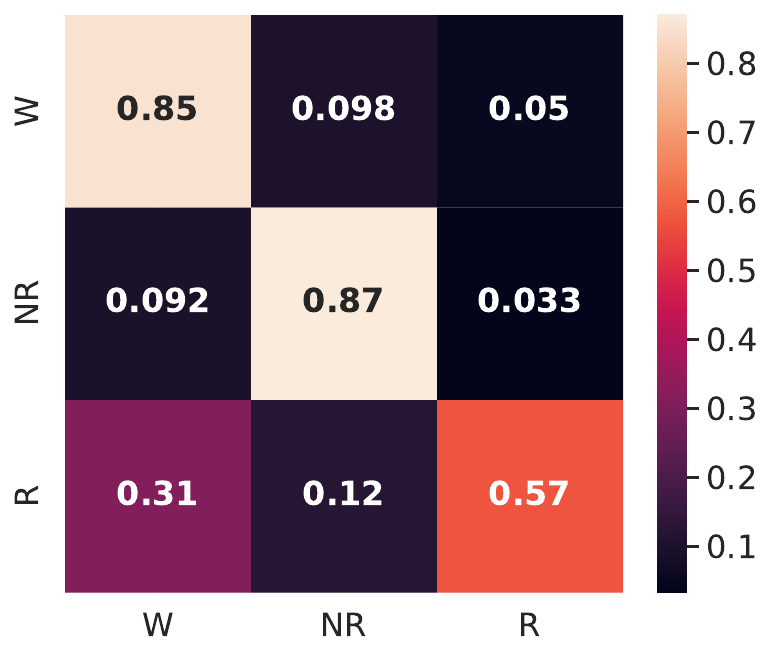}\\
    \centering{a)~LG-Sleep}
    \end{minipage}
    \begin{minipage}[c]{0.49\linewidth}
    \includegraphics[width=0.75\linewidth]{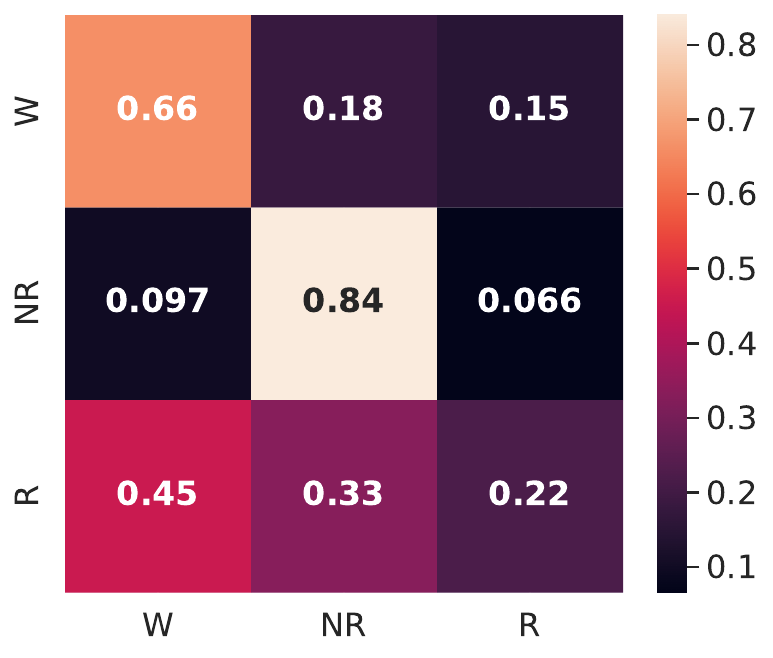}\\
    \centering{b) CNN}
    \end{minipage}
     \begin{minipage}[c]{0.49\linewidth}
    \includegraphics[width=0.75\linewidth]{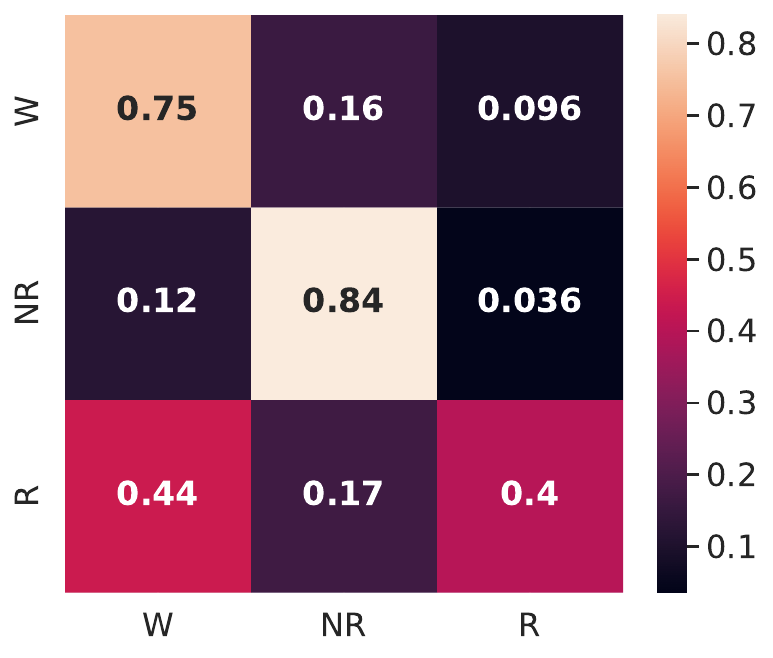}\\
    \centering{c) LSTM}
    \end{minipage}
     \begin{minipage}[c]{0.49\linewidth}
    \includegraphics[width=0.75\linewidth]{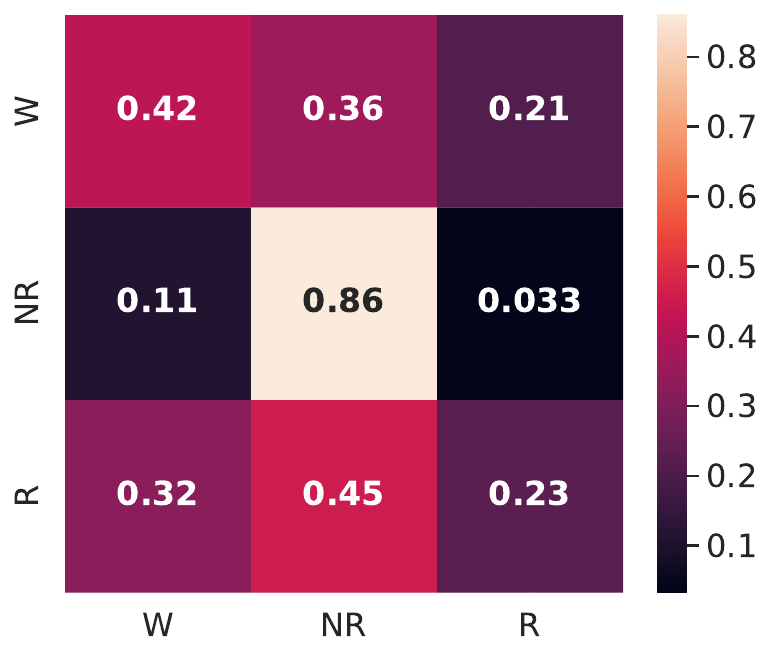}\\
    \centering{d) FC}
    \end{minipage}
    
    \caption{Confusion matrices of the proposed method and model backbones. W, NR, and R denote wake, NREM, and REM sleep stages.}
    
    \label{fig: confusion}
\end{figure}
\subsubsection{Ablation Study}
Ablation experiments were conducted to investigate the contribution of individual components or modules of the model to its overall performance. Specifically, we performed ablation experiments to evaluate the effectiveness of the proposed LG-Sleep method for sleep stage classification compared to three other methods: CNN, LSTM, and fully connected (FC). The ablation study whose results we present in Table~\ref{tab: ablation} involves a comparison of our approach with methods corresponding to subblocks of our approach, which are themselves deep learning methods. The architectures and the number of layers and parameters of these methods are the same as the corresponding subblocks in our approach, as shown in Fig.~\ref{fig: overall} and Table~\ref{tab:detailsofmodel}.

To demonstrate the methods' performance on each class, the confusion matrix for all methods is shown in Fig.~\ref{fig: confusion}. Considering REM as the most challenging class due to the lower number of samples, LG-Sleep correctly classified this class $57\%$ of the time, outperforming the other baselines. The ablation experiments provide insights into the contribution of different components and help evaluate the effectiveness of the proposed method.
\begin{table}[t]
    \centering
    \caption{Average of the classification and F$1$ score results on the test set when the number of labeled training data samples are limited.}
    \vspace{-0.3 cm}
    \begin{tabular}[t]{c c c}
    \hline
\textbf{Metric} & {$N_l=25\%N$} &  {$N_l=50\%N$}\\
        \hline
    &$N_l=3457$&$N_l=6914$ \\
Accuracy &$0.67\pm0.11$ &$0.76\pm0.06$\\ 
 F1-score & $0.50\pm0.08$& $0.63\pm0.05$\\ 
    \hline
    \end{tabular}
    \label{limitted}
\end{table}
\subsubsection{Limited Number of Training Samples}
Here, we discuss the robustness of the models while the number of labeled samples is limited. We evaluated the proposed model's performance on a test set when the number of labeled training data samples, $N_l$, is limited compared to total training samples, $N$. Table~\ref{limitted} shows the results for two different scenarios: $N_l$ is $25\%$ and $50\%$ of the total training data samples. The results indicate that using $50\%$ of labeled training samples the proposed model performs reasonably by achieving an average accuracy of $0.76\pm0.06$. Even with only $25\%$ of labeled training data, the performance is $0.67\pm0.11$ which shows LG-Sleep performs reasonably well even with limited labeled training data.
\subsubsection{Comparison With State-Of-The-Art}
The performance comparison of LG-Sleep with existing approaches for mice sleep scoring is presented in Table~\ref{tab:SoA}. All methods were re-implemented in Python using TensorFlow based on the provided descriptions. For~\cite{grieger2021automated} and \cite{yamabe2019mc}, input data samples are considered as 10-second vectors, while for~\cite{alsolai2022employing}, the input data format is the same as LG-Sleep. The remaining training setup remains consistent with the previously explained configuration. Additionally, we report results for a scenario with limited labeled training samples to highlight the impact of our proposed loss function. In~\cite{grieger2021automated}, a CNN-based deep neural network is employed, while~\cite{alsolai2022employing} introduces an LSTM-based architecture. Furthermore,~\cite{yamabe2019mc} utilizes a combination of CNN and LSTM without considering local and global temporal dependencies separately. Our results indicate that LG-Sleep achieves the highest performance. A major drawback of existing methods is their low F1-score, reflecting poor performance with limited sleep stage samples. LG-Sleep overcomes this by utilizing both short and long time steps in feature extraction and by employing a novel loss function and autoencoder-decoder layout, enhancing performance even with limited labeled samples. 

We also applied a t-test to demonstrate that the results are significant when $N=N_l$. Using a $p-value$ threshold of 0.05 as the significance level, our results show that the performance differences between LG-Sleep and the other approaches are statistically significant.
\begin{table}[t]
\centering
\caption{Comparison results with existing state-of-the-art.}
\vspace{-0.3 cm}
\begin{tabular}[t]{ c c c |c c }
\hline
 \multicolumn{1}{c}{Method}&\multicolumn{2}{c}{$N_l=N$} & \multicolumn{2}{c}{$N_l=50\%N$} \\
\hline
\hline
&\multicolumn{1}{c}{Accuracy}&\multicolumn{1}{c|}{F1-score} & \multicolumn{1}{c}{Accuracy} &  \multicolumn{1}{c}{F1-score} \\
\hline
Greiger~\textit{et al.} \cite{grieger2021automated}&$0.71\pm0.11$&$0.51\pm0.08$&$0.67\pm0.18$&$0.41\pm0.14$\\
Alsolai~\textit{et al.} \cite{alsolai2022employing}&$0.71\pm0.03$&$0.35\pm0.07$&$0.62\pm0.07$&$0.25\pm0.02$\\
Yamabe~\textit{et al.} \cite{yamabe2019mc}&$0.68\pm0.06$&$0.58\pm0.15$&$0.67\pm0.06$&$0.44\pm0.10$\\
{\bf Proposed LG-SLeep}&${\bf 0.85\pm0.09}$ & ${\bf 0.75\pm0.10}$&${\bf 0.76\pm0.06}$&${\bf 0.63\pm0.05}$\\
\hline
\end{tabular}
\label{tab:SoA}
\end{table}

\subsubsection{Limitations} LG-Sleep considers both local and global temporal dependencies, and the encoder-decoder architecture helps maintain performance even when the number of labeled training samples is limited. Although the results on LG-Sleep suggest that the method could be practical for wildtype mice sleep-scoring, it still suffers from some limitations. The main limitation is that training the DL is costly. While deep learning algorithms achieve promising accuracy, they require a sufficient number of training samples from various classes. Based on confusion matrices, LG-Sleep encounters difficulties in detecting REM due to the limited number of samples. In future, data augmentation could be used during the training phase.

\section{Conclusion}
\label{sec:conclu}
This paper presents LG-SLEEP, a novel deep learning model for classifying sleep stages using EEG signals. The model undergoes a two-phase training process: initially optimizing both MSE and classification losses to balance class weights, and then focusing on classification loss alone. LG-SLEEP achieves high accuracy ($0.85$) and an F1-score of $0.75$ in sleep stage classification. It also performs well with limited labeled training data, demonstrating its robustness. Overall, LG-SLEEP outperforms existing algorithms in sleep scoring.

\bibliographystyle{IEEEbib}
\bibliography{sn-bibliography}
\end{document}